\documentclass[10pt]{scrartcl}

\usepackage{makeidx}
\usepackage{graphicx}
\usepackage{amssymb}
\usepackage{amsmath}
\usepackage{amscd}
\usepackage{tikz}
\usetikzlibrary{matrix,arrows}
\usepackage{multicol}

\newtheorem{theorem}{Theorem}

\newtheorem{example}{Example}

\newcommand{\qed}{\hspace*{\fill} $\blacksquare$}

\newcommand{\commentout}[1]{}

\newcommand{\N}{\mathbb{N}}                    
\newcommand{\R}{\mathbb{R}}                    

\newcommand{\abs}[1]{\mathop{\left\lvert #1 \right\rvert}} 
\newcommand{\args}[1]{\mathop{\left( #1 \right)}} 
\newcommand{\inner}[1]{\mathop{\left\langle #1 \right\rangle}}
\newcommand{\norm}[1]{\mathop{\left\lVert #1 \right\rVert}}
\newcommand{\cbrace}[1]{\mathop{\left\{ #1 \right\}}}

\newcommand{\argsS}[2]{\mathop{\left( #1 \right)#2}} 

\newcommand{\normS}[2]{\mathop{\left\lVert #1 \right\rVert#2}}

\renewcommand{\S}[1]{{\mathcal{#1}}}           
\def\vec#1{\mathchoice{\mbox{\boldmath$\displaystyle#1$}}
{\mbox{\boldmath$\textstyle#1$}}
{\mbox{\boldmath$\scriptstyle#1$}}
{\mbox{\boldmath$\scriptscriptstyle#1$}}}

\renewenvironment{cases}{%
\left\{\begin{array}{c@{\quad : \quad}l}}%
{%
\end{array}\right.}

\newcommand{\atab}[1]{\hspace*{#1em}}

\newcounter{algorithm_counter}
\newenvironment{algorithm}[1]{
\refstepcounter{algorithm_counter}
\setlength{\parindent}{0\parindent}
\vspace{2ex}
\begin{minipage}{\textwidth}
\rule{\textwidth}{5\arrayrulewidth}\\
\begin{footnotesize}
{\bf Algorithm \arabic{algorithm_counter}} (#1) \\
\rule[+1.5ex]{\textwidth}{\arrayrulewidth}

\vspace{-1.5ex}

}%
{
\\[-1.5ex]
\rule{\textwidth}{\arrayrulewidth}
\end{footnotesize}
\end{minipage}
\setlength{\parindent}{\parindent}
}

{
\\[-1.5ex]
\rule{\textwidth}{\arrayrulewidth}
\end{footnotesize}
\end{minipage}
\setlength{\parindent}{\parindent}
}

\usepackage{hyperref}
\usepackage{xcolor,colortbl}

\begin{document}

\title{Generalized Gradient Learning on Time Series under Elastic Transformations}

\author{Brijnesh J.~Jain \\
       Technische Universit\"at Berlin, Germany\\
       e-mail: brijnesh.jain@gmail.com}
            
\date{}

\maketitle

\begin{abstract}
The majority of machine learning algorithms assumes that objects are represented as vectors. But often the objects we want to learn on are more naturally represented by other data structures such as sequences and time series. For these representations many standard learning algorithms are unavailable. We generalize gradient-based learning algorithms to time series under dynamic time warping. To this end, we introduce elastic functions, which extend functions on time series to matrix spaces. Necessary conditions are presented under which generalized gradient learning on time series is consistent.  We indicate how results carry over to arbitrary elastic distance functions and to sequences consisting of symbolic elements. Specifically, four linear classifiers are extended to time series under dynamic time warping and applied to benchmark datasets. Results indicate that generalized gradient learning via elastic functions have the potential to complement the state-of-the-art in statistical pattern recognition on time series.
\end{abstract}


\section{Introduction}

Statistical pattern recognition on time series finds many applications in diverse domains such as speech recognition, medical signal analysis, and recognition of gestures \cite{Fu2011,Geurts2001}. A challenge in learning on time series consists in filtering out the effects of shifts and distortions in time. A common and widely applied approach to address invariance of shifts and distortions are elastic transformations such as dynamic time warping (DTW). Following this approach amounts in learning on time series spaces equipped with an elastic proximity measure. 

In comparison to Euclidean spaces, mathematical concepts such as the derivative of a function and a well-defined addition under elastic transformations are unknown in time series spaces. Therefore gradient-based algorithms can not be directly applied to time series. The weak mathematical structure of time series spaces bears two consequences: (a) there are only few learning algorithms that directly operate on time series under elastic transformation; and (b) simple methods like the nearest neighbor classifier together with the DTW distance belong to the state-of-the-art and are reported to be \emph{exceptionally difficult to beat}  \cite{Bastista2011,Lines2014,Xi2006}. 
 
To advance the state-of-the-art in learning on time series, first adaptive methods have been proposed. They mainly devise or apply different measures of central tendency of a set of time series under dynamic time warping \cite{Kruskal1983,Rabiner1979,Rabiner1980,Petitjean2011}. The individual approaches reported in the literature are k-means \cite{Hautamaki2008,Niennattrakul2007a,Niennattrakul2007b,Petitjean2014,Wilpon1985}, self-organizing maps \cite{Somervuo1999}, and learning vector quantization \cite{Somervuo1999}. These methods have been formulated in a problem-solving manner without a unifying theme. Consequently, there is no link to a mathematical theory that allows us to (1) place existing adaptive methods in a proper context, (2) derive adaptive methods on time series other than those based on a concept of mean, and (3) prove convergence of adaptive methods to solutions that satisfy necessary conditions of optimality. 

Here we propose generalized gradient methods on time series spaces that combine the advantages of gradient information and elastic transformation such that the above issues (1)--(3) are resolved. The key idea behind this approach is the concept of elastic function.  Elastic functions extend functions on Euclidean spaces to time series spaces such that elastic transformations are preserved. Then learning on time series amounts in minimizing piecewise smooth risk functionals using generalized gradient methods proposed by \cite{Ermoliev1998,Norkin1986}. Specifically, we investigate elastic versions of logistic regression, (margin) perceptron learning, and linear support vector machine (SVM) for time series under dynamic time warping. We derive update rules and present different convergence results, in particular an elastic version of the perceptron convergence theorem. Though the main treatment focuses on univariate time series under DTW, we also show under which conditions the theory also holds for multivariate time series and sequences with non-numerical elements under arbitrary elastic transformations. 

We tested the four elastic linear classifiers to all two-class problems of the UCR time series benchmark dataset \cite{Keogh2011}. The results show that elastic linear classifiers on time series behave similarly to linear classifiers on vectors. Furthermore, our findings indicate that generalized gradient learning on time series spaces have the potential to complement the state-of-the-art in statistical pattern recognition on time series, because the simplest elastic methods are already competitive with the best available methods.

The paper is organized as follows: Section 2 introduces background material. Section 3 proposes elastic functions, generalized gradient learning on sequence data, and elastic linear classifiers. In Section 4, we relate the proposed approach to previous approaches on averaging a set of time series. Section 5 presents and discusses experiments. Finally, Section 6 concludes with a summary of the main results and an outlook for further research.

\section{Background}

This section introduces basic material. Section \ref{sec:dtw} defines the DTW distance, Section \ref{sec:learning} presents the problem of learning from examples, and Section \ref{sec:piecewise-smooth} introduces piecewise smooth functions.

\subsection{Dynamic Time Warping Distance}\label{sec:dtw}

By $[n]$ we denote the set $\cbrace{1, \ldots, n}$ for some $n \in \N$. A time series of length $n$ is an ordered sequence $\vec{x} = (x_1, \ldots, x_n)$ with features $x_i \in \R$ sampled at discrete points of time $i \in [n]$. 

To define the DTW distance between time series $\vec{x}$ and $\vec{y}$ of length $n$ and $m$, resp., we construct a grid $\S{G} = [n] \times [m]$. A warping path in grid $\S{G}$ is a sequence $\phi = (\vec{t}_1, \ldots, \vec{t}_p)$ consisting of points $\vec{t}_k = \args{i_{k}, j_{k}} \in \S{G}$ such that 
\begin{enumerate}
\item $\vec{t}_1 = (1, 1)$ and $\vec{t}_p = (n, m)$ \hfill (boundary conditions)
\item $\vec{t}_{k+1} - \vec{t}_k \in \cbrace{(1,0), (0,1), (1,1)}$ \hfill (warping conditions)
\end{enumerate}
for all $1 \leq k < p$. 

A warping path $\phi$ defines an alignment between sequences $\vec{x}$ and $\vec{y}$ by assigning elements $x_{i}$ of sequence $\vec{x}$ to elements $y_{j}$ of sequence $\vec{y}$ for every point $(i, j) \in \phi$. The boundary condition enforces that the first and last element of both time series are assigned to one another accordingly. The warping condition summarizes what is known as the monotonicity and continuity condition. The monotonicity condition demands that the points of a warping path are in strict ascending lexicographic order. The continuity condition defines the maximum step size between two successive points in a path. 

The cost of aligning $\vec{x} = (x_1, \ldots, x_n)$ and $\vec{y} = (y_1, \ldots, y_m)$ along a warping path $\phi$ is defined by
\[
d_{\phi}(\vec{x}, \vec{y}) = \sum_{(i,j)\in \phi} c\argsS{x_{i}, y_{j}},
\]
where $c(x_{i}, y_{j})$ is the local transformation cost of aligning features $x_i$ and $y_j$. Unless otherwise stated, we assume that the local transformation costs are given by 
$c\argsS{x_{i}, y_{j}} = \argsS{x_i - y_j}{^2}$. Then the distance function
\[
d(\vec{x},\vec{y}) = \min_{\phi} \, \sqrt{d_{\phi}(\vec{x}, \vec{y})},
\]
is the dynamic time warping (DTW) distance between $\vec{x}$ and $\vec{y}$, where the minimum is taken over all warping paths in $\S{G}$.

\subsection{The Problem of Learning}\label{sec:learning} 
We consider learning from examples as the problem of minimizing a risk functional. To present the main ideas, it is sufficient to focus on supervised learning. 

Consider an input space $\S{X}$ and output space $\S{Y}$. The problem of supervised learning is to estimate an unknown function $f_*:\S{X} \rightarrow \S{Y}$ on the basis of a training set
\[
\S{D} = \cbrace{\args{x_1, y_1}, \ldots, \args{x_N, y_N}} \subseteq \S{X} \times \S{Y},
\]
where the examples $(x_i, y_i) \in \S{X} \times \S{Y}$ are drawn independent and identically distributed according to a joint probability distribution $P(x, y)$ on $\S{X} \times \S{Y}$. 

To measure how well a function $f:\S{X} \rightarrow \S{Y}$ predicts output values $y$ from $x$, we introduce the risk
\begin{align*}
R[f] = \int_{\S{X}\times\S{Y}} \ell(y, f(x)) \,dP(x, y),
\end{align*}
where $\ell:\S{Y} \times \S{Y} \rightarrow \R_+$ is a loss function that quantifies the cost of predicting $f(x)$ when the true output value is $y$. 

The goal of learning is to find a function $f:\S{X} \rightarrow \S{Y}$ that minimizes the risk. The problem is that we can not directly compute the risk of $f$, because the probability distribution $P(x, y)$ is unknown. But we can use the training examples to estimate the risk of $f$ by the empirical risk
\[
R_N[f] = \frac{1}{N}\sum_{i=1}^N \ell(y_i, f(x_i)).
\]
The empirical risk minimization principle suggests to approximate the unknown function $f_*$ by a function 
\[
f_N = \arg\min_{f \in \S{F}} R_N[f]
\]
that minimizes the empirical risk over a fixed hypothesis space $\S{F} \subset \S{Y^X}$ of functions $f:\S{X} \rightarrow \S{Y}$.

Under appropriate conditions on $\S{X}$, $\S{Y}$, and $\S{F}$, the empirical risk minimization principle is justified in the following sense: 
(1) a minimizer $f_N$ of the empirical risk exists, though it may not be unique; and (2) the risk $R[f_N]$ converges in probability to the risk $R[f_*]$ of the best but unknown function $f_*$ when the number $N$ of training examples goes to infinity. 

\subsection{Piecewise Smooth Functions}\label{sec:piecewise-smooth}

A function $f:\S{X} \rightarrow \R$ defined on a Euclidean space $\S{X}$ is piecewise smooth, if $f$ is continuous and there is a finite collection of continuously differentiable functions 
$\S{R}(f) = \cbrace{f_i: \S{X} \rightarrow \R \,:\, i \in \S{I}}$ indexed by the set $\S{I}$ such that 
\[
f(x) \in \cbrace{f_i(x) \,:\, i \in \S{I}}
\]
for all $x \in \S{X}$. We call the collection $\S{R}(f)$ a representation for $f$. A function $f_i \in \S{R}(f)$ satisfying
$f_i(x) = f(x)$ is an active function of $f$ at $x$. The set $\S{A}(f, x) = \cbrace{i \in \S{I} \,:\, f_i(x) = f(x)}$  is the active index set of $f$ at $x$. By 
\[
\partial f(x) = \cbrace{\nabla f_{i}(x) \,:\, i \in \S{A}(f, x)}
\]
we denote the set of active gradients $\nabla f_{i}(x)$ of active function $f_i$ at $x$. Active gradients are directional derivatives of $f$. At differentiable points $x$ the set of active gradients is of the form $\partial f(x) = \cbrace{\nabla f(x)}$.

Piecewise smooth functions are closed under composition, scalar multiplication, finite sums, pointwise max- and min-operations. In particular, the max- and min-operations of a finite collection of differentiable functions allow us to construct piecewise smooth functions. Piecewise functions $f$ are non-differentiable on a set of Lebesgue measure zero, that is $f$ is differentiable almost everywhere.

\section{Generalized Gradient Learning on Time Series Spaces}

This section generalizes gradient-based learning to time series spaces under elastic transformations. We first present the basic idea of the proposed approach in Section \ref{subsec:idea}. Then Section \ref{sec:elastic} introduces the new concept of elastic functions. Based on this concept, Section \ref{sec:supervised} describes supervised generalized gradient learning on time series. As an example, Section \ref{sec:linear} introduces elastic linear classifiers. In Section \ref{sec:mean}, we consider unsupervised generalized gradient learning. Section \ref{subsec:note} sketches consistency results. Finally, Section \ref{sec:general} generalizes the proposed approach to other elastic proximity functions and arbitrary sequence data. 

\subsection{The Basic Idea}\label{subsec:idea}

This section presents the basic idea of generalized gradient learning on time series. For this we assume that $\S{F_X}$ is a hypothesis space consisting of functions $F:\S{X} \rightarrow \R$ defined on some Euclidean space $\S{X}$. For example,  $\S{F_X}$ consists of all linear functions on $\S{X}$. First we show how to generalize functions $F \in\S{F_X}$  defined on Euclidean spaces to functions $f: \S{T} \rightarrow \R$ on time series such that elastic transformations are preserved. The resulting functions $f$ are called elastic. Then we turn the focus on learning an unknown elastic function over the new hypothesis space $\S{F_T}$ of elastic functions obtained from $\S{F_X}$.

We define elastic functions $f:\S{T}\rightarrow \R$ on time series as a pullback of a function $F\in \S{F_X}$ by an embedding $\mu: \S{T} \rightarrow \S{X}$, that is
$f(\vec{x}) = F(\mu(\vec{x}))$ for all time series $\vec{x} \in \S{T}$. 

\begin{figure}[t]
\centering
 \includegraphics[width=0.75\textwidth]{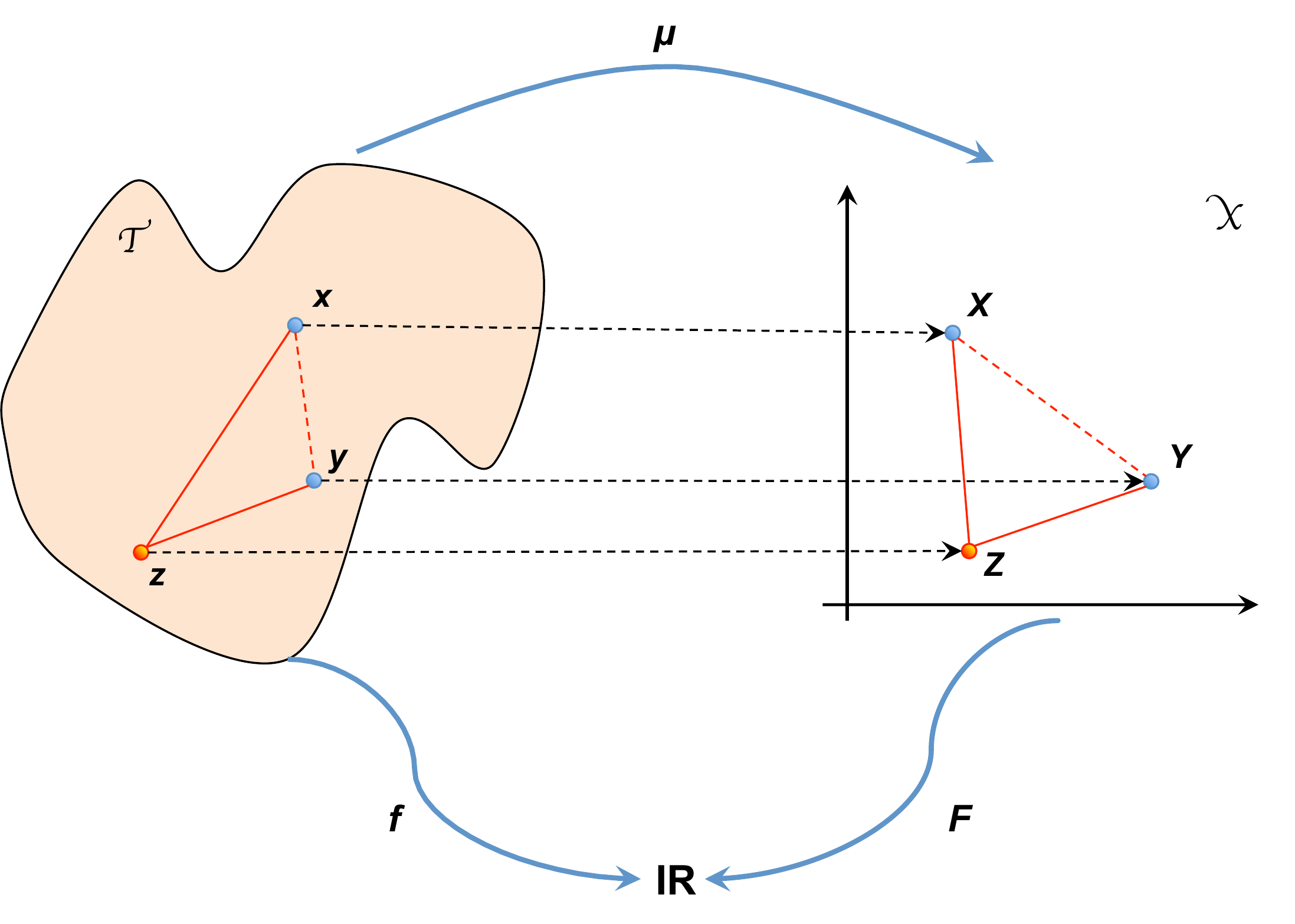}
\caption{Illustration of elastic function $f:\S{T} \rightarrow \R$ of a function $F:\S{X} \rightarrow \R$. The map $\mu = \mu_{\vec{z}}$ embeds time series space $\S{T}$ into the Euclidean space $\S{X}$. Corresponding solid red lines indicate that distances between respective endpoints are preserved by $\mu$. Corresponding dashed red lines show that distances between respective endpoints are not preserved. The diagram commutes, that is $f(\vec{x}) = F(\mu(\vec{x}))$ is a pullback of $F$ by $\mu$.}
\label{fig:elastic} 
\end{figure}

In principle any injective map $\mu$ can be used. Here, we are interested in embeddings that preserve elastic transformations. For this, we select a problem-dependent base time series $\vec{z} \in \S{T}$. Then we define an embedding $\mu_{\vec{z}}: \S{T} \rightarrow \S{X}$ that is isometric with respect to $\vec{z}$, that is 
\[
d(\vec{x}, \vec{z}) = \norm{\mu_{\vec{z}}(\vec{x})-\mu_{\vec{z}}(\vec{z})}
\]
for all $\vec{x} \in \S{T}$. It is important to note that an embedding $\mu_{\vec{z}}$ is distance preserving with respect to $\vec{z}$, only. In general, we will have $d(\vec{x}, \vec{y}) \leq \norm{\mu_{\vec{z}}(\vec{x})-\mu_{\vec{z}}(\vec{y})}$ showing that an embedding $\mu_{\vec{z}}$ will be an expansion of the time series space. This form of a restricted isometry turns out to be sufficient for our purposes. We call the pullback $f = F\circ \mu$ of $F$ by $\mu$ elastic, if embedding $\mu$ preserves elastic distances with respect to some base time series. Figure \ref{fig:elastic} illustrates the concept of elastic function.

Next we show how to learn an unknown elastic function by risk minimization over the hypothesis space $\S{F_T}$ consisting of pullbacks of functions from $\S{F_X}$ by $\mu$. For this we assume that $\Theta_{\S{T}}$ is a set of parameters and the hypothesis space  $\S{F_T}$ consists of functions $f_{\vec{\theta}}$ with parameter $\vec{\theta} \in \Theta_{\S{T}}$. To convey the basic idea, we consider the simple case that the parameter set is of the form $\Theta_{\S{T}} = \S{T}$. Then the goal is to minimize a risk functional 
\begin{align}\label{eq:problem:01}
\quad \min_{\vec{\theta} \in \S{T}} R[\vec{\theta}]
\end{align}
as a function of $\vec{\theta}\in \S{T}$. We cast problem \eqref{eq:problem:01} to the equivalent problem  
\begin{align}\label{eq:problem:02}
\quad \min_{\vec{\theta} \in \S{T}} R[\mu(\vec{\theta})],
\end{align}
Observe that the risk functional of problem \eqref{eq:problem:02} is a function of elements $\mu(\vec{\theta})$ from the Euclidean space $\S{X}$. Since problem \eqref{eq:problem:02} is analytically difficult to handle, we consider the relaxed problem 
\begin{align}\label{eq:problem:03}
\quad \min_{\vec{\Theta} \in \S{X}} R[\vec{\Theta}],
\end{align}
where the minimum is taken over the whole set $\S{X}$, whereas problem \eqref{eq:problem:02} minimizes over the subset $\mu(\S{T}) \subset \S{X}$. The relaxed problem \eqref{eq:problem:03} is not only analytically more tractable but also learns a model from a larger hypothesis space and may therefore provide better asymptotical solutions, but may require more training data to reach acceptable test error rates \cite{Vapnik1994}.  

\subsection{Elastic Functions}\label{sec:elastic}

\begin{figure}[t]
\centering
 \includegraphics[width=0.95\textwidth]{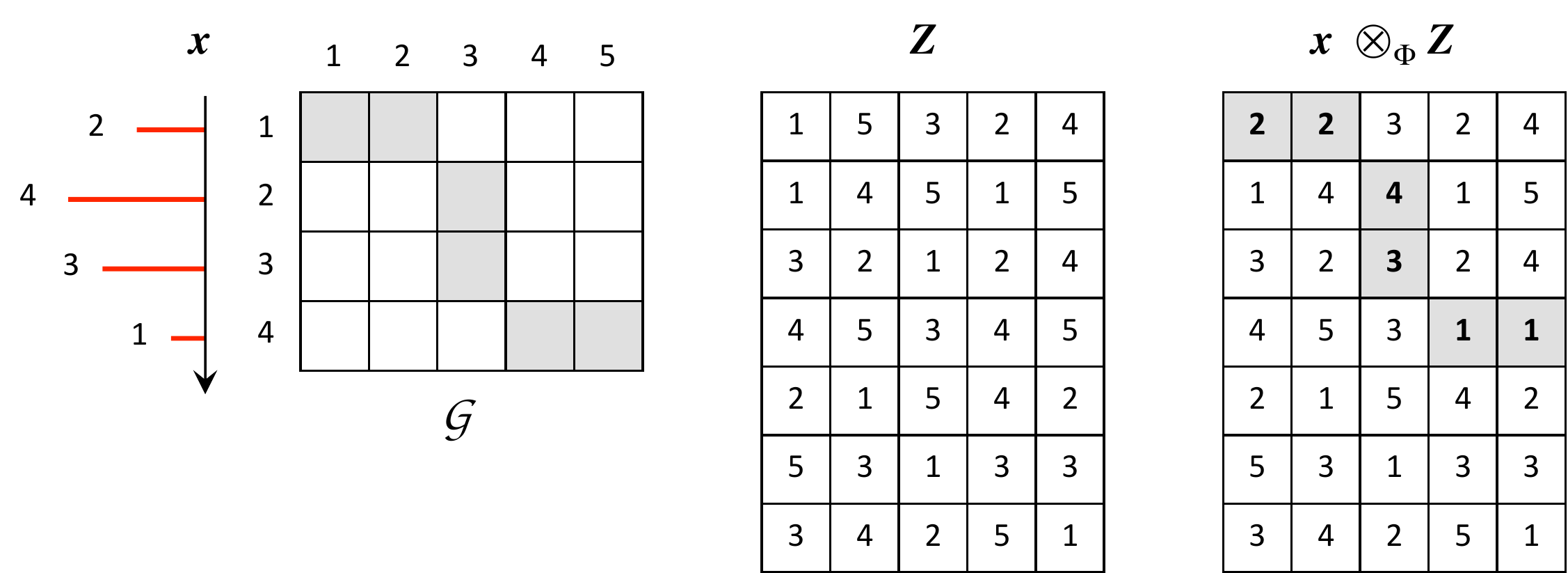}
\caption{Embedding of time series $\vec{x} = (2, 4, 3, 1)$ into matrix $\vec{Z}$ along warping path $\phi$. From left to right: Time series $\vec{x}$, grid $\S{G}$ with highlighted warping path $\phi$, matrix $\vec{Z}$, and matrix $\vec{x} \otimes_\phi \vec{Z}$ obtained after embedding $\vec{x}$ into $\vec{Z}$ along $\phi$. We assume that the length of the longest time series in the training set is $n = 7$. Therefore the matrix $\vec{Z}$ has $n=7$ rows. The number $m$ of columns of $\vec{Z}$ is a problem dependent parameter and set to $m=5$ in this example. Since time series $\vec{x}$ has length $k = 4$, the grid $\S{G} = [k]\times[m]$ containing all feasible warping paths consists of $4$ rows and $5$ columns. Grids $\S{G}$ vary only in the number $k$ of rows in accordance with the length $k \leq n$ of the time series to be embedded, but always have $m$ columns.}
\label{fig:embedding} 
\end{figure}

This section formally introduces the concept of elastic function, which generalize functions on matrix spaces $\S{X} = \R^{n \times m}$ to time series spaces. The matrix space $\S{X}$ is the Euclidean space of all real ($n \times m$)-matrices with inner product 
\[
\inner{\vec{X}, \vec{Y}} = \sum_{i,j} x_{ij}\cdot y_{ij}.
\]
for all $\vec{X}, \vec{Y} \in \S{X}$. The inner product induces the Euclidean norm
\[
\norm{\vec{X}} = \sqrt{\inner{\vec{X}, \vec{X}}}
\]
also known as the Frobenius norm.\footnote{We call $\norm{\vec{X}}$ Euclidean norm to emphasize that we regard $\S{X}$ as a Euclidean space.} The dimension $n \times m$ of $\S{X}$ has the following meaning: the number $n$ of rows refers to the maximum length of all time series from the training set $\S{D}$. The number $m$ of columns is a problem dependent  parameter, called \emph{elasticity} henceforth. A larger number $m$ of columns admits higher elasticity and vice versa.

We first define an embedding from time series into the Euclidean space $\S{X}$. We embed time series into a matrix from $\S{X}$ along a warping path as illustrated in Figure \ref{fig:embedding}. Suppose that $\vec{x} = (x_1, \ldots, x_k)$ is a time series of length $k \leq n$. By $\S{P}(\vec{x})$ we denote the set of all warping paths in the grid $\S{G} = [k] \times [m]$ defined by the length $k$ of $\vec{x}$ and elasticity $m$. An elastic embedding of time series $\vec{x}$ into matrix $\vec{Z} = (z_{ij})$ along warping path $\phi \in \S{P}(\vec{x})$ is a matrix $\vec{x} \otimes_{\phi} \!\vec{Z} = (x_{ij})$ with elements
\[
x_{ij} = \begin{cases}
x_i & (i,j) \in \phi\\
z_{ij} & \text{otherwise}
\end{cases}.
\]
Suppose that $F: \S{X} \rightarrow \R$ is a function defined on the Euclidean space $\S{X}$. An elastic function of $F$ based on matrix $\vec{Z}$ is a function $f: \S{T} \rightarrow \R$ with the following property: for every time series $\vec{x} \in \S{T}$ there is a warping path $\phi \in \S{P}(\vec{x})$ such that 
\[
f(\vec{x}) = F(\vec{x} \otimes_\phi \vec{Z}).
\]
The representation set and active set of  $f$ at $\vec{x}$ are of the form 
\begin{align*}
\S{R}(f,\vec{x})& = \cbrace{F(\vec{x} \otimes_\phi \vec{Z}) \,:\, \phi \in \S{P}(\vec{x})}\\
\S{A}(f, \vec{x})&= \cbrace{\phi \in \S{P}(\vec{x}) \,:\, f(\vec{x}) = F(\vec{x} \otimes_\phi \vec{Z})}.
\end{align*}

The definition of elastic function corresponds to the properties described in Section \ref{subsec:idea} and in Figure \ref{fig:elastic}. To see this, we define an embedding $\mu_{\vec{Z}}: \S{T} \rightarrow \S{X}$ that first selects for every time series $\vec{x}$ an active warping path $\phi \in \S{A}(f, \vec{x})$ and then maps $\vec{x}$ to the matrix $\mu_{\vec{Z}}(\vec{x}) = \vec{x} \otimes_{\phi} \vec{Z}$. Then we have $F(\mu_{\vec{Z}}(\vec{x})) = f(\vec{x})$ for all $\vec{x} \in \S{T}$. Suppose that the rows of matrix $\vec{Z}$ are all equal to $\vec{z}$. Then $\mu_{\vec{z}} = \mu_{\vec{Z}}$ is isometric with respect to $\vec{z}$.

\medskip

\begin{figure}[t]
\centering
 \includegraphics[width=0.95\textwidth]{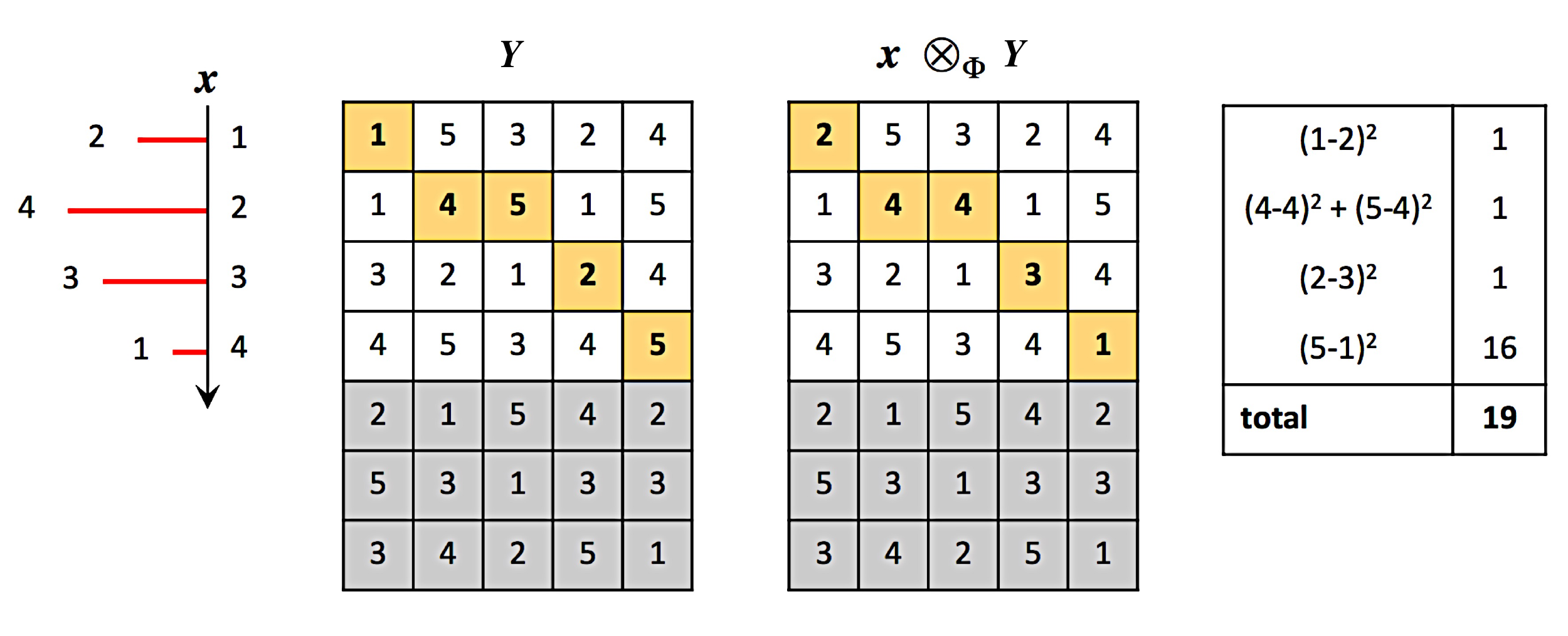}
\caption{Elastic Euclidean distance $\delta_{\vec{Y}}(\vec{x})$. From left to right: time series $\vec{x} = (2, 4, 3, 1)$, matrix $\vec{Y}$, matrix $\vec{x} \otimes_\phi \!\vec{Y}$ obtained by embedding $\vec{x}$ into matrix $\vec{Y}$ along optimal warping path $\phi$, and distance computation by aggregating the local costs giving $\delta_{\vec{Y}}(\vec{x}) = \sqrt{19}$. The optimal path is highlighted in orange in $\vec{Y}$ and in $\vec{x} \otimes_\phi \!\vec{Y}$. Gray shaded areas in both matrices refer to parts that  are not used, because the length $k = 4$ of $\vec{x}$ is less than $n=7$. Since $\vec{x}$ is embedded into $\vec{Y}$ only elements lying on the path $\phi$ contribute to the distance. All other local cost between elements of $\vec{Y}$ and $\vec{x} \otimes_\phi \!\vec{Y}$ are zero.}
\label{fig:elastic_euclidean} 
\end{figure}

Next, we consider examples of elastic functions. The first two examples are fundamental for extending a broad class of gradient-based learning algorithms to time series spaces.

\begin{example}[Elastic Euclidean Distance]\label{ex:wf1}
Let $\vec{Y} \in \S{X}$. Consider the function 
\[
D_{\vec{Y}} : \S{X} \rightarrow \R_+, \quad \vec{X} \;\mapsto\; \norm{\vec{X} - \vec{Y}}
\]
Then
\[
\delta_{\vec{Y}} : \S{T} \rightarrow \R_+, \quad \vec{x} \;\mapsto \; \min_{\phi \in \S{P}(\vec{x})}\; \norm{\vec{x} \otimes_\phi \!\vec{Y} \,-\, \vec{Y}},
\]
is an elastic function of $D_{\vec{Y}}$. To see this, observe that from 
\[
\delta_{\vec{Y}}(\vec{x}) = \min_{\phi \in \S{P}(\vec{x})} \norm{\vec{x} \otimes_\phi \! \vec{Y} - \vec{Y}} = \min_{\phi \in \S{P}(\vec{x})} D_{\vec{Y}}(\vec{x} \otimes_\phi \! \vec{Y})
\]
follows $\delta_{\vec{Y}}(\vec{x}) \in \S{R}(\delta_{\vec{Y}}, \vec{x}) = \cbrace{D_{\vec{Y}}(\vec{x} \otimes_\phi \! \vec{Y})\,:\, \phi \in \S{P}(\vec{x})}$.  See Figure \ref{fig:elastic_euclidean} for an illustration. We call $\delta_{\vec{Y}}$ \emph{elastic Euclidean distance} with parameter $\vec{Y}$. 
\qed
\end{example}

\begin{example}[Elastic Inner Product]\label{ex:wf2}
Let $\vec{W} \in \S{X}$. Consider the function 
\[
S_{\vec{W}} : \S{X} \rightarrow \R, \quad \vec{X} \;\mapsto\; \inner{\vec{X}, \vec{W}}
\]
Then the function 
\[
\sigma_{\vec{W}} : \S{T} \rightarrow \R, \quad \vec{x} \; \mapsto \; \max_{\phi \in \S{P}(\vec{x})}\inner{\vec{x} \otimes_\phi \vec{0}, \,\vec{W}},
\]
is an elastic function of $S_{\vec{W}}$, called \emph{elastic inner product} with parameter $\vec{W}$. 
\qed
\end{example}

The elastic Euclidean distance and elastic inner product are elastic proximities closely related to the DTW distance, where the elastic Euclidean distance generalizes the DTW distance. The time and space complexity of both elastic proximities are $O(n m)$. If no optimal warping path is required, space complexity can be reduced to $O(\max(n, m))$. To see this, we refer to Algorithm \ref{alg:elaInnerProduct}. To obtain an optimal warping path, we can trace-back along the score matrix $\vec{S}$ in the usual way. The procedure in Algorithm \ref{alg:elaInnerProduct} applies exactly the same dynamic programming scheme as the one for the standard DTW distance and therefore has the same time and space complexity.

Observe that both elastic proximities embed time series into different matrices. Elastic Euclidean distances embed time series into the parameter matrix and elastic inner products always embed time series into the zero-matrix $\vec{0}$.

\begin{figure}[t]
\begin{algorithm}{Elastic Inner Product}\label{alg:elaInnerProduct}
\atab{1} \textbf{Input:} \\
\atab{2} -- time series $\vec{x} = (x_1, \ldots, x_k)$ with $k \leq n$\\
\atab{2} -- elasticity $m$\\
\atab{2} -- weight matrix $\vec{W} = (w_{ij}) \in \R^{n \times m}$\\[1ex]
\atab{1} \textbf{Procedure}:\\
\atab{2} Let $\vec{S} = (s_{ij}) \in \R^{k \times m}$ be the initial score matrix  \\
\atab{2} $s_{11} \leftarrow x_1 w_{11}$\\
\atab{2} \textbf{for} $i = 2$ \textbf{to} $k$ \textbf{do}\\
\atab{4} $s_{i1} \leftarrow s_{i-1,1} + x_i w_{i1}$\\[1ex]
\atab{2} \textbf{for} $j = 2$ \textbf{to} $m$ \textbf{do}\\
\atab{4} $s_{1j} \leftarrow s_{1,j-1} + x_1 w_{1j}$\\[1ex]
\atab{2} \textbf{for} $i = 2$ \textbf{to} $k$ \textbf{do}\\
\atab{4}	\textbf{for} $j = 2$ \textbf{to} $m$ \textbf{do}\\
\atab{6} $s_{ij} = x_i w_{ij} + \max\cbrace{s_{i-1,j}, s_{i, j-1}, s_{i-1, j-1}}$\\[1ex]
\atab{1} \textbf{Return:} \\
\atab{2} -- $\sigma_{\vec{W}}(\vec{x}) = s_{km}$
\end{algorithm}

\vspace{-0.3cm}

\textbf{Remark:} This algorithm can also be used to compute elastic Euclidean distances. For this, replace all products $x_i w_{ij}$ by squared costs $\argsS{x_i - w_{ij}}{^2}$ and the max-operation by a min-operation.
\smallskip
\hrule
\end{figure}

\begin{example}[Elastic Linear Function]\label{ex:wf3}
Let $\Theta = \S{X} \times \R$ be a set of parameters and let $\vec{\theta} = (\vec{W}, b) \in \Theta$ be a parameter. 
Consider the linear function 
\[
F_{\vec{\theta}} : \S{X} \rightarrow \R, \quad \vec{X} \;\mapsto\; b + S_{\vec{W}}(\vec{X}) = b + \inner{\vec{X}, \vec{W}},
\]
where $\vec{W}$ is the weight matrix and $b$ is the bias. The function 
\[
f_{\vec{\theta}}: \S{T} \rightarrow \R, \quad \vec{x} \;\mapsto\; b + \sigma_{\vec{W}}(\vec{x}),
\]
is an elastic function of $F_{\vec{\theta}}$, called \emph{elastic linear function}.
\qed
\end{example}

\begin{example}[Single-Layer Neural Network]\label{ex:wf4}
Let $\Theta = \S{X}^r \times \R^{2r+1}$ be a set of parameters. Consider the function
\[
f_{\vec{\theta}} : \S{T} \rightarrow \R, \quad \vec{x} \;\mapsto\; b + \sum_{i=1}^r w_i \,\alpha\!\args{f_i(\vec{x})},
\]
where $\alpha(z)$ is a sigmoid function, $f_i = f_{\vec{\theta}_i}$ are elastic linear functions with parameters $\vec{\theta}_i = \args{\vec{W}_{\!i}, b_i}$, and 
$\vec{\theta} = \args{\vec{\theta}_1, \ldots, \vec{\theta}_r, w_1, \ldots, w_r, b}$. The function $f_{\vec{\theta}}$ implements an elastic neural network for time series with $r$ sigmoid units in the hidden layer and a single linear unit in the output layer.  
\qed
\end{example}

\subsection{Supervised Generalized Gradient Learning}\label{sec:supervised}

This section introduces a generic scheme of generalized gradient learning for time series under dynamic time warping. 

\medskip

Let $\Theta = \S{X}^r \times \R^s$ be a set of parameters. Consider a hypothesis space $\S{F}$ of functions $f_{\vec{\theta}}:\S{T} \rightarrow \S{Y}$ with parameter 
$\vec{\theta} = \args{\vec{W}_{\!1}, \ldots, \vec{W}_{\!r}, \vec{b}} \in \Theta$. Suppose that $\S{D} = \cbrace{\args{\vec{x}_1, y_1}, \ldots, \args{\vec{x}_N}, y_N} \subseteq \S{T} \times \S{Y}$ is a training set. According to the empirical risk minimization principle, the goal is to minimize 
\[
R_N[\vec{\theta}] = R_N[f_{\vec{\theta}}] = \sum_{i=1}^N \ell(y, f_{\vec{\theta}}(\vec{x}))
\]
as a function of $\vec{\theta}$. Since $R_N$ is a function of $\vec{\theta}$, we rewrite the loss by interchanging the role of argument $\vec{z} = (\vec{x}, y)$  and parameter $\vec{\theta}$ such that 
\begin{align}\label{eq:loss-general-case}
\ell_{\vec{z}} : \Theta \rightarrow \R, \quad \vec{\theta} \;\mapsto \;\ell(y, f_{\vec{\theta}}(\vec{x})).
\end{align}
We assume that the loss $\ell_{\vec{z}}$ is piecewise smooth with representation set
\[
\S{R}(\ell_{\vec{z}}) = \cbrace{\ell_{\vec{\Phi}}: \Theta \rightarrow \R \,:\, \vec{\Phi} = (\phi_1, \ldots, \phi_r)\in \S{P}^r(\vec{x})}
\]
indexed by $r$-tuples of warping paths from $\S{P}(\vec{x})$. The gradient $\nabla \ell_{\vec{\Phi}}$ of an active function $\ell_{\vec{\Phi}}$ at $\vec{\theta}$ is given by
\[
\nabla \ell_{\vec{\Phi}} = \args{\frac{\partial \ell_{\vec{\Phi}}}{\partial \vec{W}_{\!1}}, \ldots, \frac{\partial \ell_{\vec{\Phi}}}{\partial \vec{W}_{\!r}}, \frac{\partial \ell_{\vec{\Phi}}}{\partial \vec{b}}},
\]
where $\partial \ell_{\vec{\Phi}}/\partial \vec{\theta}_i$ denotes the partial derivative of $\ell_{\vec{\Phi}}$ with respect to $\vec{\theta}_i$.
The incremental update rule of the generalized gradient method is of the form
\begin{align}
\label{eq:update-general-case-1}
\vec{W}_{\!i}^{t+1} &= \vec{W}_{\!i}^t - \eta^t \cdot \frac{\partial}{\partial \vec{W}_{\!i}^t} \,\ell_{\vec{\Phi}}\!\args{\vec{\theta}^t}\\
\label{eq:update-general-case-2}
\vec{b}^{t+1} &= \vec{b}^t - \eta^t \cdot \frac{\partial}{\partial \vec{b}^t} \,\ell_{\vec{\Phi}}\!\args{\vec{\theta}^t}
\end{align}
for all $i \in [r]$. Section \ref{subsec:note} discusses consistency of variants of update rule \eqref{eq:update-general-case-1} and \eqref{eq:update-general-case-2}.

\subsection{Elastic Linear Classifiers}\label{sec:linear}

Let $\S{Y} = \cbrace{\pm 1}$ be the output space consisting of two class labels. An elastic linear classifier is a function of the form 
\begin{align}\label{eq:linear-classifier}
h_{\vec{\theta}} :\S{T} \rightarrow \S{Y}, \quad \vec{x} \mapsto \begin{cases}
+1 & f_{\vec{\theta}}(\vec{x}) \geq 0\\
-1 & f_{\vec{\theta}}(\vec{x}) < 0
\end{cases}
\end{align}
where $f_{\vec{\theta}}(\vec{x}) = b + \sigma_{\vec{W}}(\vec{x})$ is an elastic linear function and $\vec{\theta} = (\vec{W}, b)$ summarizes the parameters. We assign a time series $\vec{x}$ to the positive class if $f_{\vec{\theta}}(\vec{x}) \geq 0$ and to the negative class otherwise.

\begin{table}[t]
\centering
\begin{tabular}{lll}
\hline
\hline
\\[-2ex]
\commentout{\multicolumn{2}{l}{\textbf{Elastic Adaline}}&$\S{Y} = \cbrace{\pm 1}$\\
\quad loss function	 		& $\ell = \frac{1}{2}\!\argsS{y - f_{\vec{\theta}}(\vec{x})}{^2}$\\[0.5ex]
\quad partial derivative		& $\partial_{\,\vec{W}} \ell = -\args{y -f_{\vec{\theta}}(\vec{x})} \cdot \,\vec{X}$\\[1ex]
}%
\multicolumn{2}{l}{\textbf{Elastic Logistic Regression}}& $\S{Y} = \cbrace{0, 1}$\\
\quad logistic function 	& $g_{\vec{\theta}}(\vec{x}) = 1/ \args{1+\exp(-f_{\vec{\theta}}(\vec{x})}$\\[0.5ex]
\quad loss function		& $\ell = -y \log(g_{\vec{\theta}}(\vec{x})) - (1-y) \log(1-g_{\vec{\theta}}(\vec{x}))$\\[0.5ex]
\quad partial derivative		& $\partial_{\,\vec{W}} \ell = -\args{y-g_{\vec{\theta}}(\vec{x}))}\cdot \,\vec{X}$\\[1ex]
\multicolumn{2}{l}{\textbf{Elastic Perceptron}}&$\S{Y} = \cbrace{\pm 1}$\\ 
\quad loss function 		& $\ell = \max\cbrace{0, -y\cdot f_{\vec{\theta}}(\vec{x})}$\\[0.5ex]
\quad partial derivative		& $\partial_{\,\vec{W}} \ell = -y\cdot \vec{X}\cdot \mathbb{I}_{\cbrace{\ell > 0}}$\\[1ex]
\multicolumn{2}{l}{\textbf{Elastic Margin Perceptron}}&$\S{Y} = \cbrace{\pm 1}$\\
\quad loss function		& $\ell =\max\cbrace{0, \xi-y\cdot f_{\vec{\theta}}(\vec{x})}$\\[0.5ex]
\quad partial derivative		& $\partial_{\,\vec{W}} \ell = -y\cdot \vec{X}\cdot \mathbb{I}_{\cbrace{\ell > 0}}$\\[1ex]
\multicolumn{2}{l}{\textbf{Elastic Linear SVM}}&$\S{Y} = \cbrace{\pm 1}$ \\
\quad loss function		& $\ell = \lambda \normS{\vec{W}}{^2} + \max\cbrace{0, 1 - y\cdot f_{\vec{\theta}}(\vec{x})}$\\[0.5ex]
\quad partial derivative		& $\partial_{\,\vec{W}} \ell = -y\cdot \vec{X} \cdot\mathbb{I}_{\cbrace{\ell > 0}}$\\[1ex]
\hline
\hline
\end{tabular}
\caption{Examples of elastic linear classifiers. By $\partial_{\,\vec{W}} \ell$ we denote a partial derivative of an active function of $\ell$ with respect to $\vec{W}$. The partial derivatives $\partial_{b} \ell$ coincide with their corresponding counterparts in vector spaces and are therefore not included. The matrix $\vec{X} = \vec{x} \otimes_\phi \vec{0}$ is obtained by embedding time series $\vec{x}$ into the zero-matrix $\vec{0}$ along active warping path $\phi$. The indicator function $\mathbb{I}_{\cbrace{z}}$ returns $1$ if the boolean expression $z$ is true and returns $0$, otherwise. The elastic perceptron is a special case of elastic margin perceptron with margin $\xi = 0$. The elastic linear SVM can be regarded as a special $L_2$-regularized elastic margin perceptron with margin $\xi = 1$.}
\label{tab:classifiers}
\end{table}

Depending on the choice of loss function $\ell(y, f_{\vec{\theta}}(\vec{x}))$, we obtain different elastic linear classifiers as shown in Table \ref{tab:classifiers}.
The loss function of elastic logistic regression is differentiable as a function of $f_{\vec{\theta}}$ and $b$, but piecewise smooth as a function of $\vec{W}$. All other loss functions are piecewise smooth as a function of $f_{\vec{\theta}}$, $b$ and $\vec{W}$. 

From the partial derivatives, we can construct the update rule of the generalized gradient method. For example, the incremental / stochastic update rule of the elastic perceptron is of the form
\begin{align}
\label{eq:ggm-elastic-perceptron-1}
\vec{W}^{t+1} &= \vec{W}^t + \eta^t \,y \vec{X}\\
\label{eq:ggm-elastic-perceptron-2}
b^{t+1} & = b^t + \eta^t \, y,
\end{align}
where $(\vec{x}, y)$ is the training example at iteration $t$, and $\vec{X} = \vec{x} \otimes_\phi \vec{0}$ with $\phi \in \S{A}(\ell,\vec{x})$ . From the factor $\mathbb{I}_{\cbrace{\ell > 0}}$ shown in Table \ref{tab:classifiers} follows that the update rule given in \eqref{eq:ggm-elastic-perceptron-1} and \eqref{eq:ggm-elastic-perceptron-2} is only applied when $\vec{x}$ is misclassified.

\medskip

\noindent
We present three convergence results. A proof is given in Appendix \ref{app:convergence-perceptron}.

\medskip 

\noindent
\emph{Convergence of the generalized gradient method.}\ The generalized gradient method for minimizing the empirical risk of an elastic linear classifier with convex loss converges to a local minimum under the assumptions of \cite{Ermoliev1998}, Theorem 4.1. 

\medskip 

\noindent
\emph{Convergence of the stochastic generalized gradient method}.
This method converges to a local minimum of the expected risk of an elastic linear classifier with convex loss under the assumptions of \cite{Ermoliev1998}, Theorem 5.1. 

\medskip 

\noindent
\emph{Elastic margin perceptron convergence theorem}.
The perceptron convergence theorem states that the perceptron algorithm with constant learning rate finds a separating hyperplane, whenever the training patterns are linearly separable. A similar result holds for the elastic margin perceptron algorithm. 

A finite training set $\S{D} \subseteq \S{T} \times \S{Y}$ is elastic-linearly separable, if there are parameters $\vec{\theta} = (\vec{W}, b)$ such that $h_{\vec{\theta}}(\vec{x}) = y$ 
for all examples $(\vec{x}, y) \in \S{D}$. We say, $\S{D}$ is elastic-linearly separable with margin $\xi > 0$ if 
\[
\min_{(\vec{x},y) \in \S{D}} \;y \args{b + \sigma(\vec{x}, \vec{W})} \geq \xi.
\]
Then the following convergence theorem holds:
\begin{theorem}[Elastic Margin Perceptron Convergence Theorem]\label{theorem:perceptron-convergence-theorem}
Suppose that $\S{D}\subseteq \S{T} \times \S{Y}$ is elastic-linearly separable with margin $\xi > 0$. Then the elastic margin perceptron algorithm with fixed learning rate $\eta$ and margin-parameter $\lambda \leq \xi$ converges to a solution $(\vec{W}, b)$ that correctly classifies the training examples from $\S{D}$ after a finite number of update steps, provided the learning rate is chosen sufficiently small. 
\end{theorem}

\subsection{Unsupervised Generalized Gradient Learning}\label{sec:mean}

Several unsupervised learning algorithms such as, for example, k-means, self-organizing maps, principal component analysis, and mixture of Gaussians are based on the concept of (weighted) mean. Once we know how to average a set of time series, extension of mean-based learning methods to time series follows the same rules as for vectors. Therefore, it is sufficient to focus on the problem of averaging a set of time series.

Suppose that $\S{D} = \cbrace{\vec{x}_1, \ldots, \vec{x}_N} \subset \S{T}$ is a set of unlabeled time series. Consider the sum of squared distances 
\begin{align}\label{eq:elastic-mean}
F(\vec{Y})= \sum_{i=1}^N \min \cbrace{\normS{\vec{x}_i \otimes_{\phi_i} \vec{Y} - \vec{Y}}{^2} \,:\, \phi_i \in \S{P}(\vec{x}_i)}.
\end{align}
A matrix $\vec{Y}_{\!*}$ that minimizes $F$ is a mean of the set $\S{D}$ and the minimum value $F_* = F(\vec{Y}_{\!*})$ is the variation of $\S{D}$. 
The update rule of the generalized gradient method is of the form
\begin{align}\label{eq:ggm-elastic-mean}
\vec{Y}^{t+1} = \vec{Y}^t - \eta^t \sum_{i=1}^N \args{\vec{X}_i - \vec{Y}^t},
\end{align}
where $\vec{X}_i = \vec{x}_i \otimes_{\phi_{i}} \vec{Y}^t$ is the matrix obtained by embedding the $i$-th training example $\vec{x}_i$ into matrix $\vec{Y}^t$ along active warping path $\phi_i$. Under the conditions of \cite{Ermoliev1998}, Theorem 4.1, the generalized gradient method for minimizing $f$ using update rule \eqref{eq:ggm-elastic-mean} is consistent in the mean and variation.

We consider the special case, when the learning rate is constant and takes the form $\eta^t = 1/N$ for all $t\geq 0$. Then update rule \eqref{eq:ggm-elastic-mean} is equivalent to 
\begin{align}\label{eq:ggm-elastic-mean-2}
\vec{Y}^{t+1} = \frac{1}{N} \sum_{i=1}^N \vec{X}_i,
\end{align}
where $\vec{X}_i$ is as in \eqref{eq:ggm-elastic-mean}.

\subsection{A Note on Convergence and Consistency}\label{subsec:note}

Gradient-based methods in statistical pattern recognition typically assume that the functions of the underlying hypothesis space is differentiable. However, many loss functions in machine learning are piecewise smooth, such as, for example, the loss of perceptron learning, k-means, and loss functions using $\ell_1$-regularization. This case has been discussed and analyzed by \cite{Bottou2004}. 

When learning in elastic spaces, hypothesis spaces consist of piecewise smooth functions, which are pullbacks of smooth functions. Since piecewise smooth functions are closed under composition, the situation is similar as in standard pattern recognition, where hypothesis spaces consist of smooth functions. What has changed is that we will have "more" non-smooth points. Nevertheless, the set of non-smooth points remains negligible in the sense that it forms a set of Lebesgue measure zero. 

Piecewise smooth functions are locally Lipschitz and therefore admit a Clarke's subdifferential $Df$ at each point \cite{Clarke1975}. A Clarke's subdifferential $Df$ is a set that contains elements, called generalized gradients. At differentiable points, the Clarke subdifferential coincides with the gradient, that is $Df(x) = \cbrace{\nabla f(x)}$. A necessary condition of optimality of $f$ at $x$ is $0 \in Df(x)$. 

Using these and other concepts from non-smooth analysis, we can construct minimization procedures that generalize gradient descent methods. In previous subsections, we presented a slightly simpler variant of the following generalized gradient method: Consider the minimization problem
\begin{align}\label{eq:ps-problem}
\min_{x \in \S{Z}} &\quad f(x),
\end{align}
where $f$ is a piecewise smooth function and $\S{Z} \subseteq \S{X}$ is a bounded convex constraint set. Let $\S{Z}_*$ denote the subset of solutions satisfying the necessary condition of optimality and $f(\S{Z}_*) = \cbrace{f(x) \,:\, x \in \S{Z}_*}$ is the set of solution values. Consider the following iterative method:
\begin{align}
\label{eq:ggm-first}
x^0 &\in \S{Z}\\
x^{t+1}&\in \Pi_{\S{Z}}\args{x^t - \eta^t \cdot g^{t}},
\end{align}
where $g^t \in Df(x^t)$ is a generalized gradient of $f$ at $x^t$, $\Pi_{\S{Z}}$ is the multi-valued projection onto $\S{Z}$ and $\eta^t$ is the learning rate satisfying the conditions
\begin{align}
\label{eq:ggm-last}
\lim_{t \to \infty} \eta^t = 0 \qquad \text{and} \qquad
\sum_{t=0}^\infty \eta^t = \infty.
\end{align}
The generalized gradient method  \eqref{eq:ggm-first}--\eqref{eq:ggm-last} minimizes a piecewise smooth function $f$ by selecting a generalized gradient, performing the usual update step, and then projects the updated point to the constraint set $\S{Z}$. If $f$ is differentiable at $x^t$, which is almost always the case, then the update amounts to selecting an  active index $i \in \S{A}(f, x)$ of $f$ at the current iterate $x^t$ and then performing gradient descent along direction $-\nabla f_i(x^t)$. 

Note that the constraint set $\S{Z}$ has been ignored in previous subsections. We introduce a sufficiently large constraint set $\S{Z}$ to ensure convergence. In a practical setting, we may ignore specifying $\S{Z}$ unless the sequence $\args{x^t}$ accidentally goes to infinity.  

Under mild additional assumptions, this procedure converges to a solution satisfying the necessary condition of optimality \cite{Ermoliev1998}, Theorem 4.1: \emph{The sequence $\args{x^t}$ generated by method \eqref{eq:ggm-first}--\eqref{eq:ggm-last} converges to the solution of problem \eqref{eq:ps-problem} in the following sense:
\begin{enumerate}
\item the limits points $\bar{x}$ of $\args{x^t}$ with minimum value $f(\bar{x})$ are contained in $\S{Z}_*$.
\item the limits points $\bar{f}$ of $\args{f(x^t)}$ are contained in $f(\S{Z}_*)$.
\end{enumerate}
}
Consistency of the stochastic generalized gradient method for minimizing the expected risk functional follows from \cite{Ermoliev1998}, Theorem 5.1, provided similar assumptions are satisfied.

\subsection{Generalizations}\label{sec:general}

This section indicates some generalizations of the concept of elastic functions.

\subsubsection{Generalization to other Elastic Distance Functions}
Elastic functions as introduced here are based on the DTW distance via embeddings along a set of feasible warping paths with squared differences as local transformation costs. The choice of distance function and local transformation cost is arbitrary. We can equally well define elastic functions based on proximities other than the DTW distance. Results on learning carry over whenever a proximity $\rho$ on time series satisfies the following sufficient conditions: (1) $\rho$ minimizes the costs over a set of feasible paths, (2) the cost of a feasible path is a piecewise smooth function as a function of the local transformation costs, and (3) the local transformation costs are piecewise smooth.

With regard to the DTW distance, these generalizations include the Euclidean distance and DTW distances with additional constraints such as the Sakoe-Chiba band \cite{Sakoe1978}. Furthermore, absolute differences as local transformation cost are feasible, because the absolute value function is piecewise smooth.

\subsubsection{Generalization to Multivariate Time Series}
A multivariate time series is an ordered sequence $\vec{x} = \args{\vec{x}_1, \ldots, \vec{x}_n}$ consisting of feature vectors $\vec{x}_i \in \R^d$. We can define the DTW distance between multivariate time series $\vec{x}$ and $\vec{y}$ as in the univariate case but replace the local transformation cost $c(x_i, y_j) = (x_i-y_j)^2$ by $c(\vec{x}_i, \vec{y}_{\!j}) = \normS{\vec{x}_i - \vec{y}_{\!j}}{^2}$.

To define elastic functions, we embed multivariate time series into the set $\S{X} = (\R^d)^{n \times m}$ of vector-valued matrices $\vec{X} = (\vec{x}_{ij})$ with elements $\vec{x}_{ij} \in \R^d$. These adjustment preserve piecewise smoothness, because the Euclidean space $\S{X}$ is a direct product of lower-dimensional Euclidean spaces. 

\subsubsection{Generalization to Sequences with Symbolic Attributes}

We consider sequences $\vec{x} = \args{x_1, \ldots, x_n}$ with attributes $x_i$ from some finite set $\S{A}$ of $d$ attributes (symbols). Since $\S{A}$ is finite, we can represent its attributes $a \in \S{A}$ by $d$-dimensional binary vectors $\vec{a} \in \cbrace{0,1}^d$, where all but one element is zero. The unique non-zero element has value one and is related to attribute $a$. In doing so, we can reduce the case of attributed sequences to the case of multivariate time series. 

We can introduce the following local transformation costs
 \[
 c(x_i, y_j) = \begin{cases}
 0 & x_i = y_j\\
 1 & x_i \neq y_j
 \end{cases}.
 \]
 More generally, we can define local transformation costs of the form 
 \[
 c(x_i, y_j) = k(x_i, x_i) - 2k(x_i, y_j) + k(y_j, y_j),
 \]
 where $k:\S{A} \times \S{A} \rightarrow \R$ is a positive-definite kernel.
Provided that the kernel is an inner product in some finite-dimensional feature space, we can reduce this generalization also to the case of multivariate time series. 

\section{Relationship to Previous Approaches}\label{sec:relation}

Previous work on adaptive methods either focus on computing or are based on a concept of (weighted) mean of a set of time series. Most of the literature is summarized in \cite{Kruskal1983,Petitjean2011,Petitjean2014,Somervuo1999}. To place those approaches into the framework of elastic functions, it is sufficient to consider the problem of computing a mean of a set of time series. 

Suppose that $\S{D} = \cbrace{\vec{x}_1, \ldots, \vec{x}_N}$ is a set of time series. A mean is any time series $\vec{y}_{\!*}$ that minimizes the sum of squared DTW distances 
\[
f(\vec{y}) = \sum_{i=1}^N d^2(\vec{x}_i, \vec{y}).
\]

\begin{figure}[t]
\begin{algorithm}{Mean Computation}\label{alg:mean}
\atab{1} \textbf{Input:} \\
\atab{2} -- sample $\S{D} = \cbrace{\vec{x}_1, \ldots, \vec{x}_N}\subseteq \S{T}$\\[1ex]
\atab{1} \textbf{Procedure}:\\
\atab{2} 1. initialize $\vec{Y} \in \S{Z}$\\
\atab{2} 2. \textbf{repeat}\\
\atab{4} 2.1. determine active warping paths $\phi_i$ that embed $\vec{x}_i$ into $\vec{Y}$ \\
\atab{4} 2.2. update $\vec{Y} \leftarrow \upsilon(\vec{Y}, \vec{x}_1, \ldots, \vec{x}_N, \phi_1, \ldots, \phi_N)$\\
\atab{4} 2.3. project $\vec{Y} \leftarrow \pi(\vec{Y})$ to $\S{Z}$\\
\atab{2} \phantom{2.} \textbf{until} convergence\\[1ex]
\atab{1} \textbf{Return:} \\
\atab{2} -- approximation $\vec{y}$ of mean
\end{algorithm}
\end{figure}

Algorithm \ref{alg:mean} outlines a unifying minimization procedure of $f$.
The set $\S{Z}$ in line 1 of the procedure consists of all matrices with $n$ identical rows, where $n$ is the maximum length of all time series from $\S{D}$. Thus, there is a one-to-one correspondence between time series from $\S{T}$ and matrices from the subset $\S{Z}$. By construction, we have $f(\vec{y}) = F(\vec{Y})$, where $\vec{Y} \in \S{Z}$ is the matrix with all rows equal to $\vec{y}$ and $F(\vec{Y})$ is as defined in eq.~\eqref{eq:elastic-mean}. 

In line 2.1, we determine active warping paths of the function $F(\vec{Y})$ that embed $\vec{x}_i$ into matrix $\vec{Y}$. By construction this step is equivalent to computing optimal warping paths for determining the DTW distance between $\vec{x}_i$ and $\vec{y}$. 
Line 2.2 updates matrix $\vec{Y}$ and line 2.3 projects the updated matrix $\vec{Y}$ to the set $\S{Z}$. The last step is equivalent to constructing a time series from a matrix. 

Previous approaches differ in the form of update rule $\upsilon$ in line 2.2 and the projection $\pi$ in line 2.3. Algorithmically, steps 2.2 and 2.3 usually form a single step in the sense that the composition $\psi = \pi \circ \upsilon$ can not as clearly decomposed in two separate processing steps as described in Algorithm \ref{alg:mean}. The choice of $\upsilon$ and $\pi$ is critical for convergence analysis. Problems arise when the map $\upsilon$ does not select a generalized gradient and the projection $\pi$ does not map a matrix from $\S{X}$ to a closest matrix from $\S{Y}$. In these cases, it may be unclear how to define necessary conditions of optimality for the function $f$. As a consequence, even if steps 2.2 and 2.3 minimize $f$, we do not know whether Algorithm \ref{alg:mean} converges to a local minimum of $f$. The same problems arise when studying the asymptotic properties of the mean as a minimizer of $f$.

The situation is different for the function $F$  defined in eq.~\eqref{eq:elastic-mean}. When minimizing $F$, the set $\S{Z}$ coincides with $\S{X}$. Since the function $F$ is piecewise smooth, the map $\upsilon$ in line 2.2 corresponds to an update step of the generalized gradient method. The projection $\pi$ in line 2.3 is the identity.  Under the conditions of \cite{Ermoliev1998}, Theorem 4.1 and Theorem 5.1 the procedure described in Algorithm \ref{alg:mean} is consistent. 

\section{Experiments}

\begin{table}[t]
\centering
\scriptsize
\begin{tabular}{lrrrc}
\hline
\hline
Dataset & \#(Train) & \#(Test) & Length & $\rho$\\
\hline
\\[-2ex]
Coffee & 28 & 28 & 286 & 0.098\\
ECG200 & 100 & 100 & 96 & 1.042\\
ECGFiveDays & 23 & 861 & 136 & 0.169\\
Gun Point & 50 & 150 & 150 & 0.333\\
ItalyPowerDemand & 67 & 1,029 & 24 & 2.792\\
Lightning 2 & 60 & 61 & 637 & 0.094\\
MoteStrain & 20 & 1,252 & 84 & 0.238\\
SonyAIBORobotSurface & 20 & 601 & 70 & 0.286\\
SonyAIBORobotSurfaceII & 27 & 953 & 65 & 0.415\\
TwoLeadECG & 23 & 1,139 & 82 & 0.280\\
Wafer & 1,000 & 6,174 & 152 & 6.579\\
Yoga & 300 & 3,000 & 426 & 0.704\\
\hline
\hline
\end{tabular}
\caption{Characteristic features of data sets for two-class classification problems. The last column shows the ratio $\rho = $ length/\#(train).}
\label{tab:data}
\end{table}

The goal of this section is to assess the performance and behavior of elastic linear classifiers.We present and discuss results from two experimental studies. The first study explores the effects of the elasticity parameter on the error rate and the second study compares the performance of different elastic linear classifiers. We considered two-class problems of the UCR time series datasets \cite{Keogh2011}. Table \ref{tab:data} summarizes characteristic features of the datasets.

\subsection{Exploring the Effects of Elasticity}

The first experimental study explores the effects of elasticity on the error rate by controlling the number of columns of the weight matrix of an elastic perceptron. 

\subsubsection{Experimental Setup.}
The elastic perceptron algorithm was applied to the Gun\_Point, ECG200, and ECGFiveDays dataset using the following setting: The dimension of the matrix space $\S{X}$ was set to $n \times m$, where $n$ is the length of the longest time series in the training set of the respective dataset. Bias and weight matrix were initialized by drawing random numbers from the uniform distribution on the interval $[-0.01, +0.01]$. The elasticity $m$ was controlled via the ratio $w = m/n$. For every $w \in \S{S}_w$  the learning rate $\eta \in \S{S}_{\eta}$ with the lowest error on the training set was selected, where the sets are of the form
\begin{align*}
\S{S}_w &= \cbrace{0, 0.05, 0.1, 0.2, 0.3, 0.4, 0.5, 0.75, 1.0, 2.0, 3.0}\\
\S{S}_{\eta} &= \cbrace{1.0, 0.7, 0.3, 0.1, 0.03, 0.01, 0.003, 0.001}.
\end{align*}
Note that the value $w = 0$ refers to $m = 1$. Thus the weight matrix collapses to a column vector and the elastic perceptron becomes the standard perceptron. To assess the generalization performance, the learned classifier was applied to the test set. The whole experiment was repeated $30$ times for every value $w$.

\subsubsection{Results and Discussion}

\begin{figure}[t]
\centering
 \includegraphics[width=0.3\textwidth]{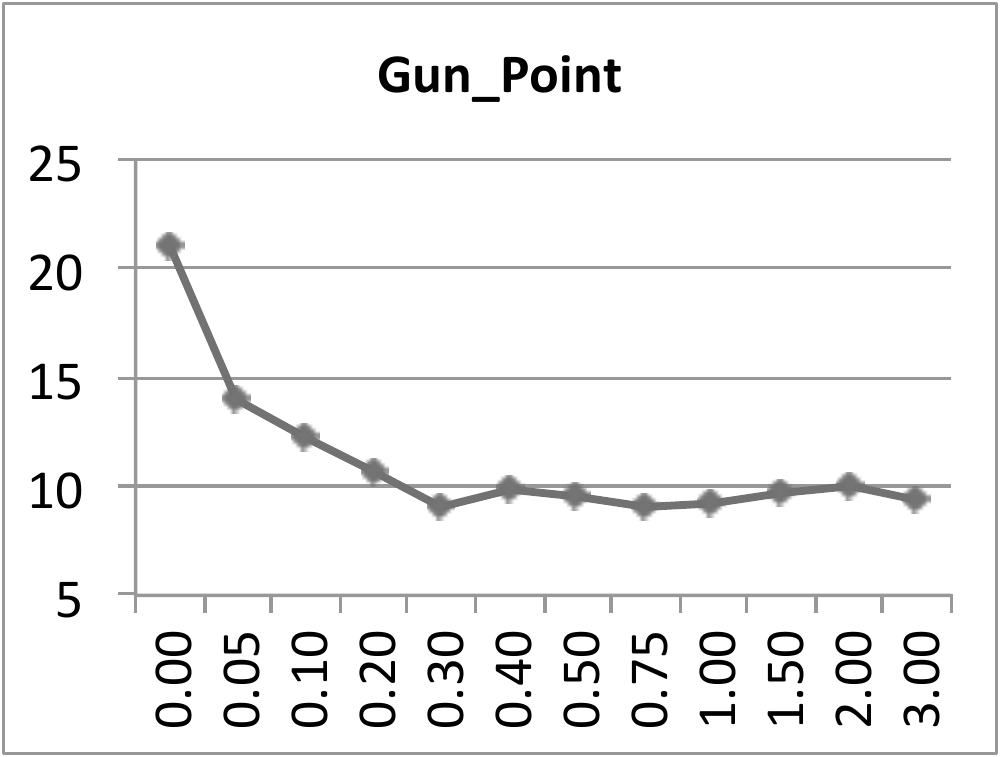} \hfill
  \includegraphics[width=0.3\textwidth]{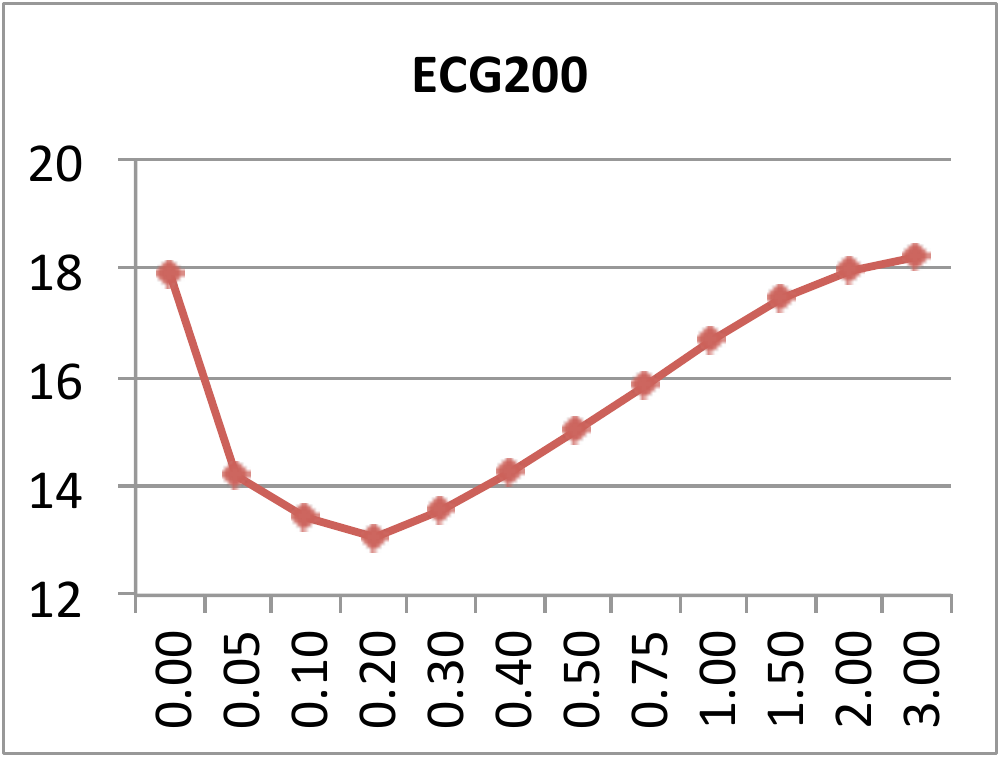} \hfill
 \includegraphics[width=0.3\textwidth]{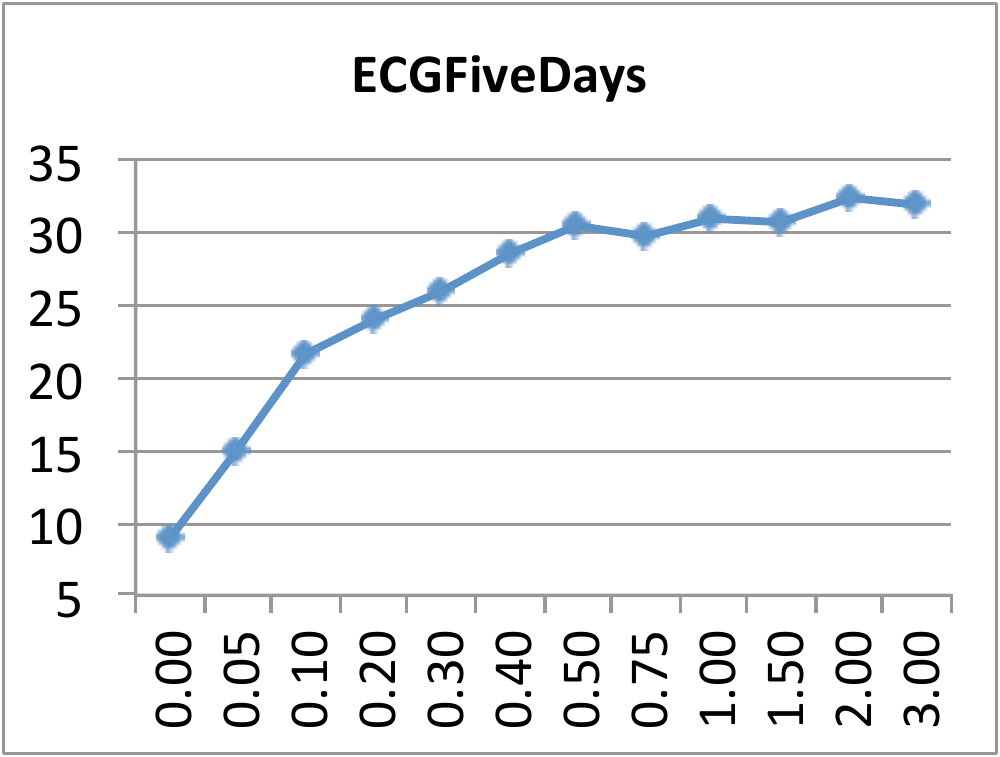}
\caption{Mean error rates of elastic perceptron on Gun\_Point, ECG200, and ECGFiveDays. Vertical axes show the mean error rates in \% averaged over 30 trials. Horizontal axes show the ratio $w = m/n$, where $m$ is the elasticity, that is the number of columns of the weight matrix and $n$ is the length of the longest time series of the respective dataset. Ratio $w = 0$ means $m = 1$ and corresponds to the standard perceptron algorithm.}
\label{fig:results} 
\end{figure}

Figure \ref{fig:results} shows the mean error rates of the elastic perceptron as a function of $w = m/n$. The error rates on the respective training sets were always zero.

One characteristic feature of the UCR datasets listed in Table \ref{tab:data} is that the number of training examples is low compared to the dimension of the time series.  This explains the low training error rates and the substantially higher test error rates. 

The three plots show typical curves also observed when applying the elastic perceptron to the other datasets listed in Table \ref{tab:data}. The most important observation to be made is that the parameter $w$ is problem-dependent and need to be selected carefully. If the training set is small and dimensionality is high, a proper choice of $w$ becomes challenging. The second observation is that in some cases, the standard perceptron algorithm ($w = 0$) may perform best as in ECGFiveDays. 
Increasing $w$ results in a classifier with larger flexibiltiy. Intuitively this means that an elastic perceptron can implement more decision boundaries the larger $w$ is. If $w$ becomes too large, the classifier becomes more prone to overfitting as indicated by the results on ECG200 and ECGFiveDays. We hypothesize that elasticity controls the capacity of an elastic linear classifier.

\subsection{Comparative Study}
This comparative study assesses the performance of elastic linear classifiers. 

\subsubsection{Experimental Setup.}
 
In this study, we used all datasets listed in Table \ref{tab:data}. The four elastic linear classifiers of Section \ref{sec:linear} were compared against different variants of the nearest neighbor (NN) classifier with DTW distance. The variants of the NN classifiers differ in the choice of prototypes. The first variant uses all training examples as prototypes (NN+ALL). The second and third variant learned one prototype per class from the training set using k-means (NN+KME) as second variant and agglomerative hierarchical clustering (NN+AHC) as third variant \cite{Petitjean2014}.

The settings of the elastic linear classifiers were as follows: The dimension of the matrix space $\S{X}$ was set to $n \times m$, where $n$ is the length of the longest time series in the training set and $m = \lceil n/10 \rceil$ is the elasticity. The elasticity $m$ was set to $10\%$ of the length $n$ for the following reasons:  First, $m$ should be small to avoid overfitting due to high dimensionality of the data and small size of the training set. Second, $m$ should be larger than one, because otherwise an elastic linear classifier reduces to a standard linear classifier. 

Bias and weight matrix were initialized by drawing random numbers from the uniform distribution on the interval $[-0.01, +0.01]$. 
Parameters were selected by $k$-fold cross validation on the training set of size $N$. We set $k = 10$ if $N > 30$ and $k = N$ otherwise. The following parameters were selected: learning rate $\eta$ for all elastic linear classifiers, margin $\xi$ for elastic margin perceptron, and regularization parameter $\lambda$ for elastic linear SVM. 
The parameters were selected from the following values
\begin{align*}
\eta \in \cbrace{2^{-10}, 2^{-9}, \ldots, 2^{0}},\!\!\!\! & &
\xi \in \cbrace{10^{-7}, 10^{-6}, \ldots, 10^1},\!\!\!\!  & &
\lambda \in \cbrace{2^{-10}, 2^{-9}, \ldots, 2^{-1}}\!.
\end{align*}
The final model was obtained by training the elastic linear classifiers on the whole training set using the optimal parameter(s). We assessed the generalization performance by applying the learned model to the test data. Since the performance of elastic linear classifiers depends on the random initialization of the bias and weight matrix, we repeated the last two steps $100$ times, using the same selected parameters in each trial.

\subsubsection{Results and Discussion.}

\definecolor{Green}{rgb}{0.7, 1, 0.7}
\definecolor{Yellow}{rgb}{1, 1, 0.7}
\definecolor{Red}{rgb}{1, 0.7, 0.7}
\begin{table}[t]
\centering
\scriptsize
\begin{tabular}{|l|rrr|rrrr|}
\hline
\hline
& \multicolumn{3}{c|}{NN + DTW} & \multicolumn{4}{c|}{Elastic linear classifiers} \\
Dataset & ALL & AHC & KME & ePERC & eLOGR & eMARG & eLSVM \\
\hline
& \multicolumn{3}{c|}{}&
\\[-2ex]
\rowcolor{Green}
Coffee & 17.9 & 25.0 & 25.0 & 4.6$^{\,\pm 3.0}$ & 4.5$^{\,\pm 2.9}$ & 4.7$^{\,\pm 3.3}$ & \textbf{3.1}$^{\,\pm 2.9}$ \\
\rowcolor{Green}
ECG200 & 23.0 & 28.0 & 28.0 & 13.6$^{\,\pm 1.8}$ & \textbf{11.8}$^{\,\pm 1.6}$ & 13.6$^{\,\pm 1.9}$ & 13.1$^{\,\pm 1.7}$ \\
\rowcolor{Green}
ECGFiveDays & 23.2 & 33.0 & 33.0 & 15.3$^{\,\pm 3.7}$ & 15.7$^{\,\pm 3.3}$ & 15.3$^{\,\pm 3.4}$ & \textbf{11.1}$^{\,\pm 3.0}$\\
\rowcolor{Green}
ItalyPowDem. & 5.0 & 21.5 & 21.5 & 3.8$^{\,\pm 1.2}$ & \textbf{3.0}$^{\,\pm 0.3}$ & 3.5$^{\,\pm 0.8}$ & \textbf{3.0}$^{\,\pm 0.3}$ \\
\rowcolor{Green}
Wafer & 2.0 & 69.5 & 69.5 & 1.3$^{\,\pm 0.3}$ & 1.2$^{\,\pm 0.2}$ & \textbf{1.0}$^{\,\pm 0.2}$ & \textbf{1.0}$^{\,\pm 0.2}$ \\
\rowcolor{Yellow}
Gun Point & 9.3 & 32.7 & 32.7 & 9.7$^{\,\pm 3.6}$ & 9.2$^{\,\pm 2.5}$ & 10.0$^{\,\pm 3.4}$ & \textbf{9.0}$^{\,\pm 2.8}$ \\
\rowcolor{Yellow}
SonyAIBO II & 27.5 & 21.6 & 21.6 & 27.0$^{\,\pm 3.8}$ & \textbf{20.2}$^{\,\pm 1.4}$ & 26.6$^{\,\pm 3.3}$ & 22.7$^{\,\pm 2.1}$ \\
\rowcolor{Red}
Lighting 2 & \textbf{13.1} & 36.1 & 36.1 & 44.2$^{\,\pm 4.2}$ & 44.1$^{\,\pm 4.4}$ & 44.6$^{\,\pm 4.4}$ & 47.6$^{\,\pm 2.9}$ \\
\rowcolor{Red}
MoteStrain & 16.5 & \textbf{13.3} & \textbf{13.3} & 17.2$^{\,\pm 2.6}$ & 16.0$^{\,\pm 2.3}$ & 17.6$^{\,\pm 2.6}$ & 15.8$^{\,\pm 2.3}$ \\
\rowcolor{Red}
SonyAIBO & \textbf{16.9} & 18.8 & 18.8 & 19.3$^{\,\pm 5.4}$ & 18.6$^{\,\pm 5.0}$ & 19.5$^{\,\pm 6.6}$ & 17.8$^{\,\pm 4.2}$ \\
\rowcolor{Red}
TwoLeadECG & \textbf{9.6} & 16.2 & 16.2 & 22.7$^{\,\pm 5.3}$ & 21.8$^{\,\pm 4.5}$ & 21.7$^{\,\pm 5.4}$ & 21.8$^{\,\pm 5.3}$ \\
\rowcolor{Red}
Yoga & \textbf{16.4} & 45.8 & 45.8 & 20.9$^{\,\pm 1.2}$ & 21.5$^{\,\pm 1.0}$ &21.1$^{\,\pm 1.1}$& 20.8$^{\,\pm 1.1}$\\
\hline
\hline
\end{tabular}
\caption{Mean error rates and standard deviation of elastic linear classifiers averaged over $100$ trials and error rates of nearest-neighbor classifiers using the DTW distance (NN+DTW). ALL: NN+DTW with all training examples as prototypes; AHC: NN+DTW with one prototype per class obtained from agglomerative hierarchical clustering with Ward linkage; KME: NN+DTW with one prototype per class obtained from k-means clustering; ePERC: elastic perceptron; eLOGR: elastic logistic regression; eMARG $\!=\!$ elastic margin perceptron; eLSVM: elastic linear SVM. Best (avg.) results are highlighted. Green rows: avg.~results of all elastic linear classifiers are better than the results of all NN classifiers. Yellow rows: results of elastic linear classifiers and NN classifiers are comparable. Red rows: avg.~results of all elastic linear classifiers are worse than the best result of an NN classifier. }
\label{tab:result-perceptron}
\end{table}

Table \ref{tab:result-perceptron} summarizes the error rates of all elastic linear (EL) classifiers and nearest neighbor (NN) classifiers. 

Comparison of EL classifiers and NN methods is motivated by the following reasons: First, NN classifiers belong to the state-of-the-art and are considered to be \emph{exceptionally difficult to beat}  \cite{Bastista2011,Lines2014,Xi2006}. Second, in Euclidean spaces linear classifiers and nearest neighbors are two simple but complementary approaches. Linear classifiers are computationally efficient, make strong assumptions about the data and therefore may yield stable but possibly inaccurate predictions. In contrast, nearest neighbor methods make very mild assumption about the data and therefore often yield accurate but possibly unstable predictions \cite{Hastie2001}.

The first key observation suggests that overall generalization performance of EL classifiers is comparable to the state-of-the-art NN classifier. This observation is supported by the same same number of green shaded rows (EL is better) and red shaded rows (NN is better) in Table \ref{tab:result-perceptron}. As reported by \cite{Lines2014}, ensemble classifiers of different elastic distance measures are assumed to be first approach that significantly outperformed the NN+ALL classifier on the UCR time series dataset. This result is not surprising, because in machine learning it is well known for a long time that ensemble classifiers often perform better than their base classifiers for reasons explained in \cite{Dietterich2000}. Since any base classifier can contribute to an ensemble classifier, it is feasible to restrict comparison to base classifiers such as the state-of-the-art NN+ALL classifier.

The second key observation indicates that EL classifiers are clearly superior to NN classifiers with one prototype per class, denoted by NN$_1$ henceforth. Evidence for this finding is provided by two results:  first, AHC and KME performed best among several prototype selection methods for NN classification \cite{Petitjean2014}; and second, error rates of EL classifiers are significantly better than those of NN+AHC and NN+KME for eight, comparable for two, and significantly worse for two datasets. 

The third key observation is that EL classifiers clearly better compromise between solution quality and computation time than NN classifiers. Findings reported by \cite{Wang2013} indicate that more prototypes may improve generalization performance of NN classifiers. At the same time, more prototypes increase computation time, though the differences will decrease for larger number of prototypes by applying certain acceleration techniques. At the extreme ends of the scale, we have NN+ALL and NN$_1$ classifiers. With respect to solution quality, the first key observation states that EL classifiers are comparable to the slowest NN classifiers using the whole training set as prototypes and clearly superior to the fastest NN classifiers using one prototype per class. To compare computational efficiency, we first consider the case without applying any acceleration techniques. We measure computational efficiency by the number of proximity calculations required to classify a single time series. This comparison is justified, because the complexity of computing a DTW distance and an elastic inner product are identical. Then EL classifiers are $p$-times faster than NN classifiers, where $p$ is the number of prototypes. Thus the fastest NN classifiers effectively have the same computational effort as EL classifiers for arbitrary multi-class problems, but they are not competitive to EL classifiers according to the second key observation. Next, we discuss computational efficiency  of both types of classifiers, when one applies acceleration techniques. For NN classifiers, two common techniques to decrease computation time are global constraints such as the Sakoe-Chiba band \cite{Sakoe1978} and diminishing the number of DTW distance calculations by applying lower bounding technique \cite{Ratanamahatana2004,Ratanamahatana2005}. Both techniques can equally well be applied to EL classifiers, where lower-bounding techniques need to be converted to upper-bounding techniques. Furthermore, EL classifiers can  additionally control the computational effort by the number $m$ of columns of the matrix space. Here $m$ was set to $10\%$ of the length $n$ of the shortest time series of the training set. The better performance of EL classifiers in comparison to NN$_1$ classifiers is notable, because the decision boundaries that can be implemented by their counterparts in the Euclidean space are both the set of all hyperplanes. We assume that EL classifiers outperform NN$_1$ classifiers, because learning prototypes by clustering minimizes a cluster criterion unrelated to the risk functional of a classification problem. Therefore the resulting prototypes may fail to discriminate the data for some problems. 

\commentout{Second, in view of Algorithm \ref{alg:mean}, the prototypes learned by AHC or KME correspond to matrices of an $n$-dimensional subspace of $\S{X} = \R^{n \times n}$. The weight matrix learned by an elastic linear classifier is from a ($n \times m$)-dimensional subspace of $\S{X}$. If $m = 1$, an EL classifier reduces to the standard linear classifier with $n$-dimensional weight vector. For $m > 1$ the dimension of the weight matrix of an EL classifier is $m\cdot n \geq 2n$. Here, the shortest time series has length $n = 24$ (ItalyPowerDemand) and the longest one has length $n = 637$ (Lightning 2). Thus, the number $m = \lceil n/10\rceil$ varies between $3$ and $64$ depending on the particular dataset. This shows that the dimension of the weight matrix of an EL classifier in this empirical study is always larger than $n$. Thus, EL classifier in this study are more flexible compared to both NN$_1$ classifiers, AHC and KME.}

The fourth key observation is that the strong assumption of elastic-linearly separable problems is appropriate for some problems in the time series classification. Error rates of elastic linear classifiers for Coffee, ItalyPowerDemand, and Wafer are below $5\%$. For these problems, the strong assumption made by EL classifiers is appropriate. For all other datasets, the high error rates of EL classifiers could be caused by two factors: first, the assumption that the data is elastic-linearly separable is inappropriate; and second, the number of training examples given the length of the time series is too low for learning (see ratio $\rho$ in Table \ref{tab:data}). Here further experiments are required. 

The fifth observation is that the different EL classifiers perform comparable with advantages for eLOGR and eLSVM. These findings correspond to similar findings for logistic regression and linear SVM in vector spaces.

To complete the comparison, we contrast the time complexities of all classifiers required for learning. NN+ALL requires no time for learning. The NN+AHC classifier learns a protoype for each class using agglomerative hierarchical clustering. Determining pairwise DTW distances is of  complexity $O(n^2 N(N-1)/2)$, where $n$ is the length of the time series and $N$ is the number of training examples. Given a pairwise distance matrix, the complexity of agglomerative clustering is  $O(N^3)$ in the general case. Efficient variants of special agglomerative methods have a complexity of $O(N^2)$. Thus, the complexity of NN+AHC is $O(n^2N^2)$ in the best and $O(n^2N^2 + N^3)$ in the general case. The NN+KME learns a protoype for each class using k-means under elastic transformations. Its time complexity is $O(2n^2 N t)$, where $t$ is the number of iterations required until termination. The time complexity for learning an EL classifier is $O(nmNt)$, where $m$ is the number of columns of the weight matrix. This shows that the time complexity for learning an EL classifier is the same as learning two prototypes by KME. However, in this setting, learning an EL classifier is about factor $20$ faster than KME, under the assumption that the number of iterations $t$ is the same for both methods. If the number $N$ of training examples is large, NN+AHC becomes prohibitively slow. In contrast, the learning procedures of NN+KME and EL classifiers can be terminated after some  pre-specified maximum number of iterations. In doing so, we trade solution quality against feasible computation time.

To summarize, the results show that elastic linear classifiers are simple and efficient methods. They rely on the strong assumption that an elastic-linear decision boundary is appropriate. Therefore, elastic linear classifiers may yield inaccurate predictions when the assumptions are biased towards oversimplification and/or when the number of training examples is too low compared to the length of the time series. These findings are in line with those of linear classifiers in Euclidean space.

\section{Conclusion}

This paper introduces generalized gradient methods for learning on time series under elastic transformations. This approach combines (a) the novel concept of elastic functions that links elastic proximities on time series to piecewise smooth functions with (b) generalized gradient methods for non-smooth optimization.
Using the proposed scheme, we (1) showed how a broad class of gradient-based learning can be applied to time series under elastic transformations, (2) derived general convergence statements that justify the generalizations, and (3) placed existing adaptive methods into proper context. 
Exemplarily, elastic logistic regression, elastic (margin) perceptron learning, and elastic linear SVM have been tested on two-class problems and compared to nearest neighbor classifiers using the DTW distance. Despite the simplicity in terms of the decision boundary and the computational efficiency, elastic linear classifiers perform convincing. There is still room for improvement by controlling elasticity and by applying different forms of regularization. 
The results indicate that adaptive methods based on elastic functions may complement the state-of-the-art in statistical pattern recognition on time series, in particular when powerful non-linear gradient-based methods such as deep learning are extended to time series under elastic transformations.

\small
\begin{appendix}

\section{Proof of Convergence Results for Elastic Linear Classifiers}\label{app:convergence-perceptron}

Since affine functions are convex and the maximum of convex functions is also convex, the elastic inner product is convex. In addition, the composition of convex functions is convex. Therefore the loss functions of elastic linear classifiers are convex. Then the first convergence results is shown in \cite{Shor1985}.

To show the two other convergence statements, we assume that $\abs{\S{D}} = N$. For each training example $(\vec{x}_i, y_i) \in \S{D}$ the loss  
\[
\ell_i(\vec{\theta}) = \ell_i\args{y_i, b + \sigma(\vec{x}_i, \vec{W})}
\]
is real-valued and convex, where $\vec{\theta} = (\vec{W}, b)$. Then there is a positive scalar $C_i$ that bounds the subdifferential of $\ell_i$ at $\vec{\theta}$ for all $i \in [N]$. Suppose that
\[
C = \max_{i = {1, \ldots, N}} C_i.
\]
Then from \cite{Nedic2001}, Prop.\ 2.2. follows that the incremental generalized gradient method converges to a local minimum. 

To show the Elastic Margin Perceptron Convergence Theorem, we assume that 
\[
E_N[\vec{\theta}] = \sum_{i=1}^N \ell_i(\vec{\theta})
\]
is the error without averaging operation, that is $E_N = N \cdot R_N$. By assumption, the training set $\S{D}$ is elastic-linearly separable. Then the minimum value $E_*$ of $E_N$ is zero.  From \cite{Nedic2001}, Prop.\ 2.1. follows
\[
\lim_{t \to \infty} E_N(\vec{\theta}^t) \leq E_* + \frac{\eta \cdot C^2}{2} = \frac{\eta \cdot C^2}{2},
\] 
where $\eta$ is the learning rate. Choosing $\eta \leq \xi/C^2$ gives
\[
\lim_{t \to \infty} E_N(\vec{\theta}^t) \leq \frac{\xi}{2}.
\] 
Since $\xi > 0$, this implies that there is a $t_0$ such that $\ell_t(\vec{\theta}^t) < \xi$ for all $t \geq t_0$. Here, $\ell_t$ refers to example $(\vec{x}_t, y_t) \in \S{D}$ presented at iteration $t$. From this follows that all training examples are classified correctly after a finite number of update steps, provided that $\lambda \leq \xi$.
\qed

\end{appendix}

\end{document}